%% file: corpipe24.tex
\pdfoutput=1

\documentclass[11pt]{article}

\usepackage[preprint]{acl}

\usepackage{times}
\usepackage{latexsym}

\usepackage[T1]{fontenc}
\usepackage[utf8]{inputenc}

\usepackage{microtype}

\usepackage{inconsolata}

\usepackage{amssymb}
\usepackage{booktabs}
\usepackage{graphicx}
\usepackage{makecell}
\usepackage{multirow}
\usepackage{xcolor}
\usepackage{xspace}

\newcommand{\CRAC}{CRAC 2024 Shared Task\xspace}
\newcommand{\CRAClong}{CRAC 2024 Shared Task on Multilingual Coreference Resolution\xspace}
\newenvironment{citemize}{\begin{list}{$\bullet$}{\topsep=.1\smallskipamount\itemsep=0pt\parsep=1pt\labelwidth=.5em}}{\end{list}}

\title{CorPipe at CRAC 2024: Predicting Zero Mentions from Raw Text}

\author{Milan Straka \\
  Charles University, Faculty of Mathematics and Physics \\
  Institute of Formal and Applied Linguistics \\
  Malostranské nám. 25, Prague, Czech Republic \\
  \texttt{straka@ufal.mff.cuni.cz}
  }

\usepackage{fancyhdr}
\fancypagestyle{officialbibref}{\fancyhf{}\fancyheadoffset[L,R]{50pt}\fancyhead[C]{This paper was published at \textbf{CRAC 2024} -- please cite the published version {\small\url{https://aclanthology.org/2024.crac-1.9}}.}\fancyfoot[C]{\normalsize\thepage}\addtolength{\topmargin}{-30pt}\addtolength{\headsep}{30pt}}
\begin{document}
\thispagestyle{officialbibref}
\maketitle
\begin{abstract}
  We present CorPipe 24, the winning entry to the \CRAClong. In this third
  iteration of the shared task, a novel objective is to also predict empty
  nodes needed for zero coreference mentions (while the empty nodes were
  given on input in previous years). This way, coreference resolution
  can be performed on raw text. We evaluate two model variants: a~two-stage
  approach (where the empty nodes are predicted first using a pretrained
  encoder model and then processed together with sentence words by another
  pretrained model) and a single-stage approach (where a single pretrained
  encoder model generates empty nodes, coreference mentions, and coreference
  links jointly). In both settings, CorPipe surpasses other participants by
  a large margin of 3.9 and 2.8 percent points, respectively. The source code
  and the trained model are available at
  {\small\url{https://github.com/ufal/crac2024-corpipe}}.
\end{abstract}

\section{Introduction}

The \CRAClong~\cite{novak-etal-2024-findings} is a third iteration
of a shared task, whose goal is to accelerate research in multilingual
coreference resolution~\citep{zabokrtsky-etal-2023-findings,zabokrtsky-etal-2022-findings}.
This year, the shared task features 21 datasets in 15 languages from the
CorefUD 1.2 collection~\citep{CorefUD1.2}.

Compared to the last year---apart from 4 new datasets in 3 languages---a novel
task is to predict the so-called \textit{empty nodes} (according to the
Universal Dependencies terminology; \citealt{nivre-etal-2020-universal}). The
empty nodes can be considered ``slots'' that can be part of coreference
mentions even if not being present on the surface level of a sentence. The
empty nodes are particularly useful in pro-drop languages (like Slavic and
Romance languages), where pronouns are sometimes dropped from a sentence when
they can be inferred, for example by verb morphology, like in the Czech example
\textit{``Řekl,~že~nepřijde''}, translated as \textit{``(He) said that (he)
won't come''}.

We present CorPipe 24, an improved version of our system submitted in last
years~\citep{straka-2023-ufal,straka-strakova-2022-ufal}. We evaluate two
variants of the system. In a two-stage variant, the empty nodes are first
predicted by a baseline system utilizing a pretrained language encoder
model;\footnote{Our implementation of the baseline system was available
to all shared task participants in case they do not want to predict the empty
nodes themselves.} then, the predicted empty nodes are, together with the input
words, processed by original CorPipe using another pretrained encoder.
In comparison, a single-stage variant employs a single pretrained encoder
model, which predicts the empty nodes, coreference mentions, and coreference links
jointly.

Our contributions are as follows:
\begin{citemize}
  \item We present the winning entry to the \CRAClong, surpassing other
    participants by a large margin of 3.9 and 2.8 percent points with
    a two-stage and a single-stage variant, respectively.
  \item We compare the two-stage and the single-stage settings, showing that
    the two-stage system outperforms the single-stage system by circa one
    percent points, both in the regular and the ensembled setting.
  \item Apart from the CorefUD 1.2, we evaluate the CorPipe performance
    also on OntoNotes~\citep{pradhan-etal-2013-towards}, a frequently used
    English dataset.
  \item The CorPipe 24 source code is available at
    {\small\url{https://github.com/ufal/crac2024-corpipe}} under an open-source
    license. The two-stage and the single-stage models are also released,
    under the CC BY-NC-SA license.
\end{citemize}

\section{Related Work}

Traditionally, coreference resolution was solved by first predicting the
coreference mentions and subsequently performing coreference linking
(clustering) of the predicted mentions. However, in recent years, the
end-to-end approach~\citep{lee-etal-2017-end,lee-etal-2018-higher,
joshi-etal-2019-bert,joshi-etal-2020-spanbert} has become more popular.
Indeed, the baseline of the CRAC 2022, 2023, and 2024 shared
tasks~\citep{prazak-etal-2021-multilingual} follow this approach, as well as
the second-best solution of CRAC 2022~\citep{prazak-konopik-2022-end}
and the third-best solution of CRAC 2023.

The end-to-end approach has been improved by \citet{kirstain-etal-2021-coreference}
not to explicitly construct the span representations, and by
\citet{dobrovolskii-2021-word} to consider only the word level, ignoring the
span level altogether during coreference linking. Simultaneously,
\citet{wu-etal-2020-corefqa} formulated coreference resolution in a question
answering setting, reaching superior results at the expense of substantially
more model predictions and additional question-answering data.

The current state-of-the-art results on
OntoNotes~\citep{pradhan-etal-2013-towards}, a frequently used English
coreference resolution dataset, are achieved by autoregressive models with
billions of parameters: \citet{liu-etal-2022-autoregressive} propose
a specialized autoregressive system, while \citet{bohnet-etal-2023-coreference}
employ a text-to-text paradigm. However, both these architectures must call the
trained model repeatedly to process a single sentence.

\section{Two-stage CorPipe}
\label{sec:two-stage}

The two-stage variant of CorPipe processes input in two steps: first, empty
nodes are predicted using the baseline system available to all shared task
participants; then, the coreference resolution is performed using CorPipe. This
approach is very similar to the last year's edition of the CRAC Shared Task,
where the empty nodes were already given on input. Therefore, the last year's version
\hbox{CorPipe~23}~\citep{straka-2023-ufal} can be used.

\subsection{Empty Nodes Baseline}
\label{sec:empty_nodes_baseline}

The baseline for predicting empty nodes generates for each empty node only
the minimum amount of information needed: the word order position defined by an input
word that the empty node should follow (the word order position determines the
position of the empty node in coreference mentions) and the dependency head
and the dependency relation of the empty node (required by the empty node
matching during evaluation); no forms or lemmas are predicted even if provided
in the training data. The baseline predicts the empty nodes
non-autoregressively, generating at most two empty nodes for every input word;
the input word becomes the dependency head of the predicted empty node.

\begin{figure}[t]
    \centering
    \includegraphics[width=\hsize]{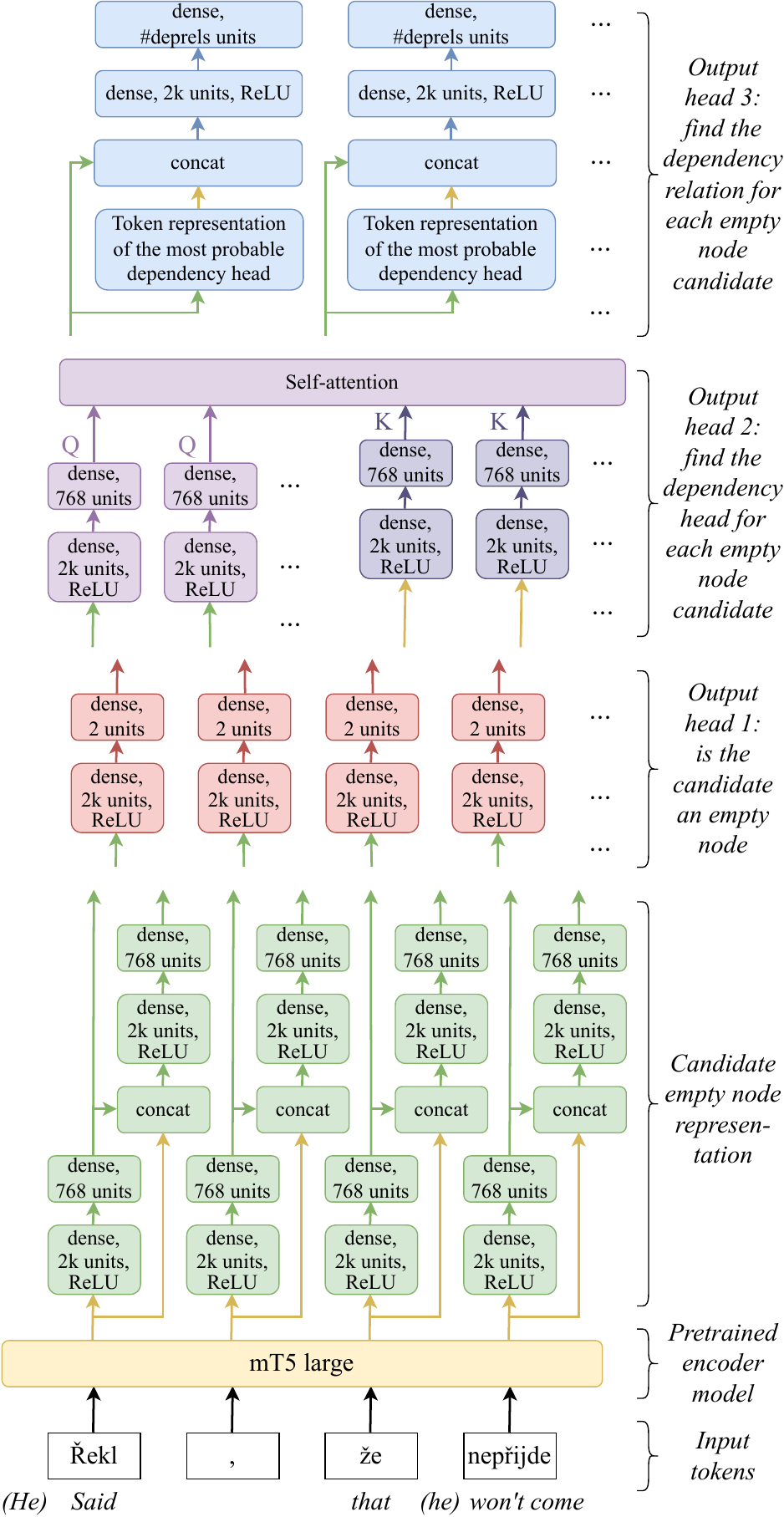}
    \caption{The system architecture of the empty node prediction baseline.
    Every ReLU activation is followed by a dropout layer layer with a dropout rate of 50\%.}
    \label{fig:baseline_architecture}
\end{figure}

The overview of the architecture is displayed in
Figure~\ref{fig:baseline_architecture}. The input words of a single sentence
are first tokenized, passed through a pretrained mT5-large
encoder~\citep{conneau-etal-2020-unsupervised}, and each input word is
represented by the embedding of its first subword. Then, the candidate for
empty nodes are generated, two per word. The first candidate is generated by
passing the input word representations through a 2k-unit dense layer with ReLU
activation, a dropout layer, and a 768-unit dense layer. The second candidate
is generated by concatenating the first candidate representation with the input
word representation and passing the result through an analogous
dense-dropout-dense module. Then, three heads are attached, each first passing
its input by a ReLU-activated 2k-unit dense layer and dropout: (1)
a classification layer deciding whether a candidate actually generates an
empty node, (2) a self-attention layer choosing the word order position (i.e.,
an input word to follow) for every candidate, and (3) a dependency relation
classification layer, which processes the candidate representation concatenated
with the representation of the word most likely according to the word-order
prediction head. Please refer to the released source code for further details.

We train a single multilingual model using the AdaFactor
optimizer~\citep{shazeer-etal-2018-adafactor} for 20 epochs, each epoch
consisting of 5\,000 batches containing 64 sentences each. The learning rate
first linearly increases from zero to the peak learning rate of 1e-5 in the
first epoch, and then decays to zero in the rest of the training according to
a cosine schedule~\citep{loshchilov-etal-2017-sgdr}. Each sentence is sampled
from the combination of all corpora containing empty nodes (see
Table~\ref{tab:empty_nodes}), proportionally to the square root of the word
size of the corresponding corpus. The model is trained for 19 hours using
a single L40 GPU with 48GB RAM.

The source code is released under the MPL license at
{\small\url{https://github.com/ufal/crac2024_zero_nodes_baseline}}, together
with the complete set of used hyperparameters. Furthermore, the trained model
is available under the CC BY-SA-NC license at
{\small\url{https://www.kaggle.com/models/ufal-mff/crac2024_zero_nodes_baseline/}}.
Finally, the development sets and the test sets of the CorefUD 1.2 datasets
with predicted empty nodes are available to all participants of the \CRAC.

\begin{table}[t]
  \input{table_empty_nodes_baseline.tex}
  \caption{Empty nodes prediction baseline performance on the development sets of CorefUD 1.2 corpora containing empty nodes. An empty node is evaluated as correct if it has the correct dependency head, dependency relation, and word order.}
  \label{tab:empty_nodes}
\end{table}

The intrinsic performance of the baseline system on the development sets of
CorefUD 1.2 is presented in Table~\ref{tab:empty_nodes}. A predicted empty
node is considered correct if it has correct dependency head, dependency
relation, and also the word order.

\subsection{Coreference Resolution}

\begin{figure*}[t]
    \centering
    \includegraphics[width=.89\hsize]{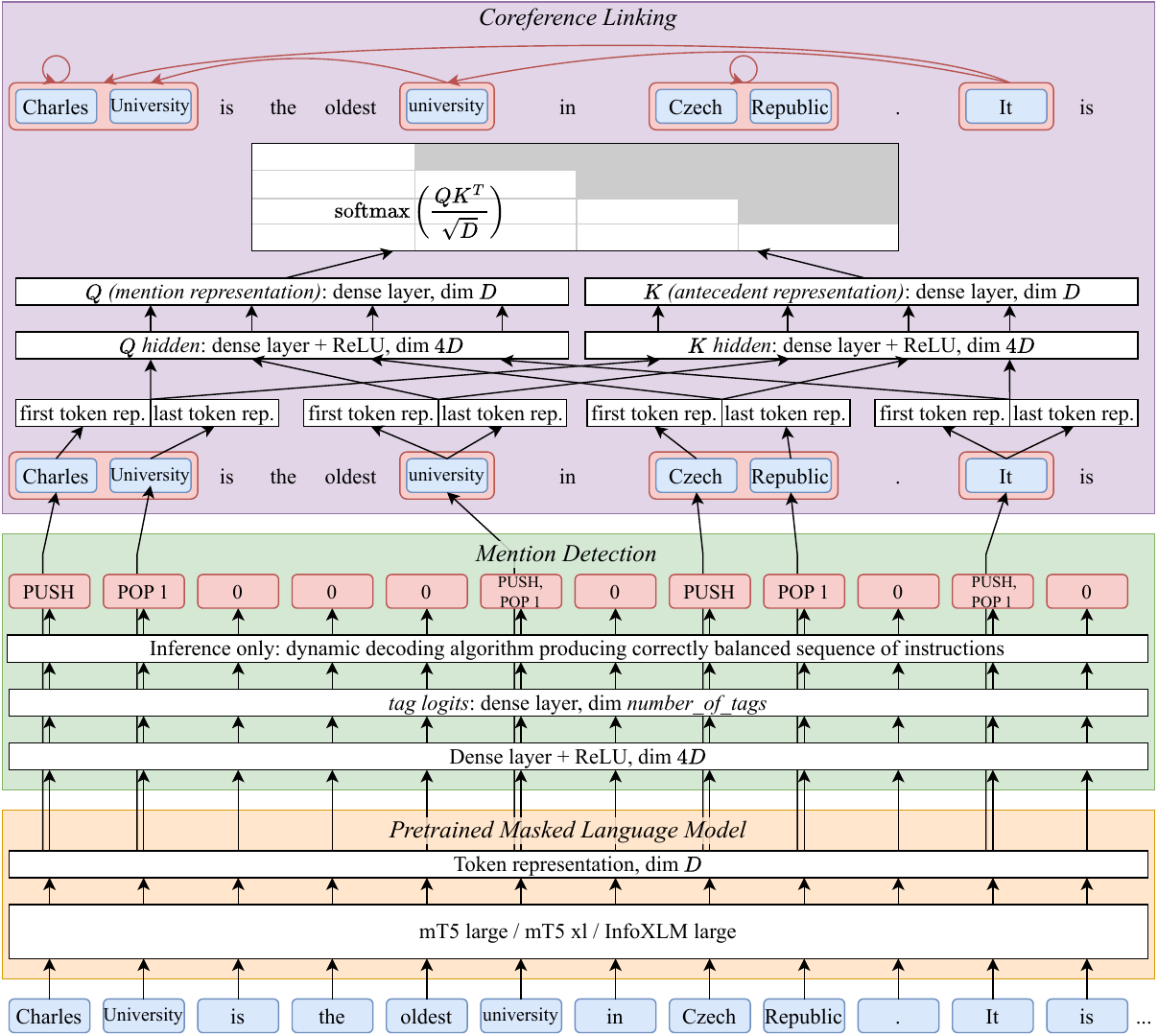}
    \caption{The CorPipe 23 model architecture introduced in~\citet{straka-2023-ufal}.}
    \label{fig:corpipe23_architecture}
\end{figure*}

With the empty nodes predicted by the baseline, we can directly employ the
CorPipe~23 from the last year of the shared task~\citep{straka-2023-ufal}. The
overview of the architecture is presented in
Figure~\ref{fig:corpipe23_architecture} and briefly described; for more
details, please refer to the original paper.

CorPipe processes the document one sentence at a time; to provide as much
context as possible, as many preceding and at most 50 following tokens are
additionally added on input, to the limit of the maximum segment size (512 or
2\,560). The words are first passed through a pretrained language encoder
model. Then, coreference mentions are predicted using an extension of BIO
encoding capable of representing possibly overlapping set of spans. Finally,
each predicted mention is represented as a concatenation of its first and last
word, and the most likely entity link (possibly to itself) of every mention is
generated using a self-attention layer.

During training, the maximum segment size is always 512; however, during
inference, we consider also larger segment size of 2\,560 for the mT5 models,
which support larger segment sizes due to their relative positional embeddings.

\subsection{Training}
\label{sec:two_stage_training}

We train the coreference resolution system analogously to the CorPipe~23
training procedure~\citep{straka-2023-ufal}. Three model variants are trained,
based on either mT5-large, mT5-xl~\citep{xue-etal-2021-mt5}, or
InfoXLM-large~\citep{chi-etal-2021-infoxlm}. For every variant, 7 multilingual
models are trained on a combination of all corpora, differing only in random
initialization. The sentences are sampled proportionally to the square
root of the word size of the corresponding corpora.

Every model is trained for 15 epochs, each epoch consisting of 10k batches.
The mT5-large and InfoXLM-large variants use the batch size of 8 and train for
14 hours on a single A100 with 40GB RAM; the mT5-xl variant employ the batch
size of 12 and train for 17 hours on 4 A100s with 40GB RAM each. The mT5 variants
are trained using the AdaFactor optimizer~\citep{shazeer-etal-2018-adafactor}
and the InfoXLM-large is trained using Adam~\citep{kingma-and-ba-2015-adam}.
The learning rate is first increased from 0 to the peak learning rate in the
first 10\% of the training and then decayed according to the cosine
schedule~\cite{loshchilov-etal-2017-sgdr}; we employ the peak learning rates
of 6e-4, 5e-4, and 2e-5 for the mT5-large, mT5-xl, and InfoXLM-large encoders,
respectively.

For each model, we keep the checkpoints after every epoch, obtaining a pool of
$3\cdot7\cdot15$ checkpoints. From this pool, we select three configurations:
(1) a single checkpoint reaching the highest development score on all the
corpora, (2) a best-performing checkpoint for every corpus according to its
development set, (3) an ensemble of 5 best-performing checkpoints for every
corpus.

\section{Single-stage CorPipe}
\label{sec:single-stage}

While the two-stage variant is full-fledged, allowing coreference mention to be
composed of any continual sequence of input words and empty nodes, it requires
two large pretrained encoders, which makes the model about twice as big and
twice as slow compared to a single model.

Therefore, we also propose a single-stage variant, with the goal of using just
a single pretrained language encoder model. For simplicity's sake, we restrict
the model in the following way: if a coreference mention contains an empty
node, the whole mention must be just this single empty node. In other words,
a coreference mention either does not contain empty nodes, or it is just
a single empty node. Note that this restriction does not decrease the score
under the head-match metric because only the mention head is used during score
computation.

With the described restriction, we no longer need to distinguish between empty
nodes and zero coreference mentions; therefore, the single-stage model predicts
only such empty nodes that are also zero coreference mentions. Finally, the
word order of an empty node is no longer needed for evaluation; as a result,
we no longer predict the word order explicitly and place the empty node after
its dependency head in the word order.

\begin{figure}[t]
    \centering
    \includegraphics[width=\hsize]{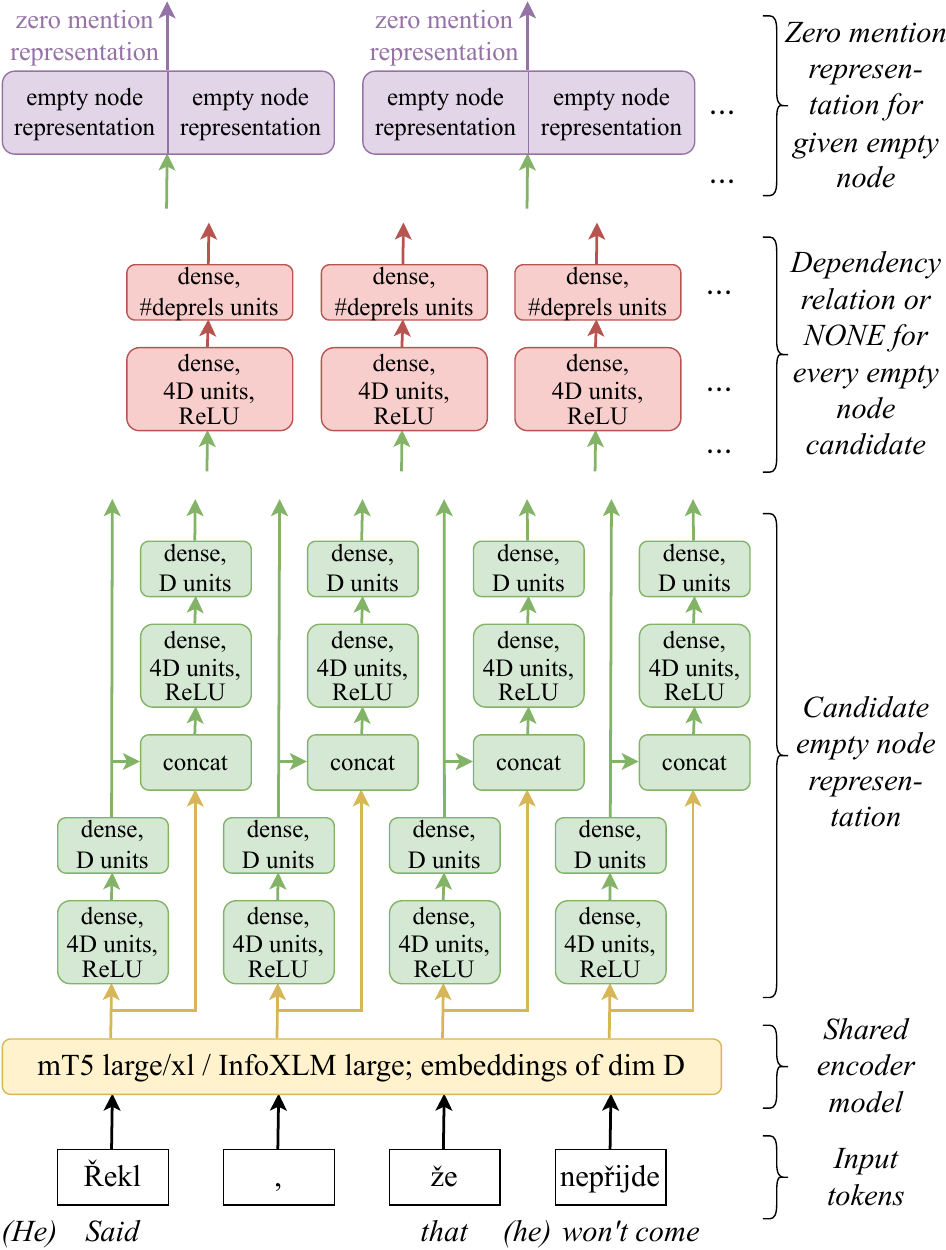}
    \caption{The changes in the CorPipe 23 architecture when empty nodes and
    zero mentions are generated jointly with mention detection and coreference
    linking.}
    \label{fig:single_stage_architecture}
\end{figure}

In Figure~\ref{fig:single_stage_architecture}, we visualize the proposed
changes to the CorPipe architecture needed to support joint empty nodes/zero
mentions prediction. Analogously to the empty nodes baseline described in
Section~\ref{sec:empty_nodes_baseline}, we start by generating two candidate
empty nodes representations from every input word representation. We then run
a classification head for every candidate, which either predicts \texttt{NONE}
when the candidate should not generate an empty node, or it predicts the
dependency relation of the generated empty node. Finally, to construct
a representation of a zero coreference mention, we concatenate the empty node
representation to itself because the empty node is both the first and the last
word of the mention. The coreference linking then proceeds as before, just
using a concatenation of surface mentions and zero mentions.

The single-stage model is trained analogously to the two-stage model. The only
differences are that (1) we pass only the input words through the pretrained
language encoder model, (2) we add the loss of the classifier predicting
dependency relation or \texttt{NONE} to the other losses (using simple
addition), and (3) we concatenate the zero mention representations to the
surface mention representations before the coreference linking step.

We closely follow the training procedure of the two-stage model described in
Section~\ref{sec:two_stage_training}. Notably, we also consider the same three
pretrained encoders, train the same number of models using the same optimizers
and learning rates, and select the same three configurations (single
best-performing checkpoint, per-corpus best checkpoint, and a per-corpus
3-model ensemble).\footnote{We only managed to use a 3-model ensemble before
the shared task deadline, while we use a 5-model ensemble for the two-stage
variant.}

\section{Shared Task Results}

\begin{table*}[t]
  \input{table_official_treebanks.tex}
  \caption{Official results of \CRAC~on the test set (CoNLL score in \%). The system $^\dagger$ is described in \citet{prazak-etal-2021-multilingual}; the rest in \citet{novak-etal-2024-findings}.}
  \label{tab:official_treebanks}
\end{table*}

\begin{table}[t]
  \input{table_official_metrics.tex}
  \caption{Official results of \CRAC~on the test set with various metrics in \%.}
  \label{tab:official_metrics}
\end{table}

In the shared task, each team was allowed to submit at most three systems. We
submitted the following configurations:
\begin{citemize}
  \item \textbf{CorPipe-single}, the large-sized single-stage model checkpoint
    achieving the best development performance across all corpora;
  \item \textbf{CorPipe}, the best-performing 3-model single-stage ensemble for
    every corpus;
  \item \textbf{CorPipe-2stage}, the best-performing 5-model two-stage ensemble
    for every corpus.
\end{citemize}
\noindent The first configuration corresponds to a real-world deployment
scenario, where a single model would be used for all corpora; the latter
configurations are the highest performing single-stage approach
(\textbf{CorPipe}, Section~\ref{sec:single-stage}) and two-stage approach
(\textbf{CorPipe-2stage}, Section~\ref{sec:two-stage}).

The official results of the shared task's primary metric are presented in
Table~\ref{tab:official_treebanks}. All our submissions outperform other
participant systems, even if \textbf{CorPipe-single} only slightly. Overall,
the ensembled single-stage variant outperforms other participants by 2.8
percent points, and the ensembled two-stage variant outperforms other
participants by 3.9 percent points.

Table~\ref{tab:official_metrics} shows the results of the submitted systems
using four metrics. Apart from the primary head-match metric, our three
submissions outperform all others also when evaluated using exact match and
with singletons. When considering partial match, the CorPipe-single is
outperformed by the system Ondfa, assumingly because it limits the predicted
mentions just to their heads, which slightly improves partial match but
severely deteriorates exact match.

\section{Ablations Experiments}

\begin{table*}[t!]
  {
  \input{table_test_results.tex}
  \caption{Ablations experiments on the CorefUD 1.2 test set (CoNLL score in \%).}
  \label{tab:test_results}
  }
  \vspace{2em}
  \input{table_ontonotes.tex}
  \caption{Comparison of CorPipe and other models on OntoNotes, using pretrained models of various size.}
  \label{tab:ontonotes}
\end{table*}

\subsection{CorefUD 1.2}

Table~\ref{tab:test_results} contains quantitative analysis of ablation
experiments on the CorefUD 1.2 test set. In Table~\ref{tab:test_results}.A, we compare the three
configurations of the single-stage model variant. Selecting the best-performing
checkpoint for every corpus increases the overall score by 1.4 percent points, while
making the model up to 21 times larger. Further addition of ensembling improves
the score by another 1.2 percent points.

The same comparison is available also for the two-stage model variant in
Table~\ref{tab:test_results}.B. We observe a similar trend of 1.2 percent
points increase for the best per-corpus checkpoint approach and further 1.4
percent points increase during ensembling.

The sections C, D, and E of Table~\ref{tab:test_results} compare the individual
checkpoint configurations of the single-stage and the two-stage models. We
observe that the effect of the two-stage model is 0.9--1.1 percent point
increase in all checkpoint configuration. We hypothesize that two factors
contribute to the better performance of the two-stage variant: first, the empty
node representation is computed by a pretrained encoder, allowing better
contextualization of the empty node representation.
Second, the mentions with empty nodes are represented in the original form,
i.e., the mentions can contain any sequence of input words and empty nodes,
while the single-stage variant represent zero mentions always by a single empty
node.

It would be interesting to evaluate the two-stage variant using the gold empty
nodes instead of predicted empty nodes to quantify the decrease of the score
caused by empty node prediction errors. Unfortunately, such an evaluation is
not supported by the shared task evaluation platform. Nevertheless,
Table~\ref{tab:test_results}.F at least shows that such a difference for the
provided baseline coreference system~\citep{prazak-etal-2021-multilingual}
is 1.4 percent points, as reported by the shared task organizers.

Finally, meaningful comparison of the shared task results between this year and
the last year is very difficult to carry out. While many corpora have changed
only marginally and the evaluation metric is the same (so the results are
reasonably comparable), other corpora have changed substantially (especially
Polish and Turkish). Even so, we provide numerical comparison of this year's
and last year's best systems in Table~\ref{tab:test_results}.G. This year's
results are slightly worse than in the last year, on average by 0.65 percent
points, but the difference is quite comparable to the effect of predicted/gold
empty nodes on the baseline system (cf. Table~\ref{tab:test_results}.F).

\subsection{OntoNotes}

To compare the performance of the CorPipe architecture to English
state-of-the-art models, we train also models on the OntoNotes
dataset~\citep{pradhan-etal-2013-towards}. The dataset does not contain any
empty nodes, so we use the last year's training setup, with the two exceptions:
we also consider pretrained English-specific encoders
T5~\citep{raffel-etal-2020-exploring} and
Flan-T5~\citep{chung-etal-2024-scaling}, and we consider larger segment size
during training (up to 1\,536 subwords).

The results are presented in Table~\ref{tab:ontonotes}. In the large-sized
setting, CorPipe outperforms all models except models utilizing additional
data~\citep{otmazgin-etal-2023-lingmess,wu-etal-2020-corefqa} and models
utilizing the word-level
approach~\citep{dobrovolskii-2021-word,doosterlinck-etal-2023-caw}.\footnote{We
are of course curious to find out how the word-level approach works on the
CorefUD dataset. Nevertheless, we hypothesize that on some of the CorefUD
corpora it might not work well because the mention heads in these corpora are
considerably less unique than in OntoNotes.}
In the xl-sized settings, our model is 0.3 percent points below the state of
the art of~\citet{liu-etal-2022-autoregressive}; notably, CorPipe outperforms
the state of the art system~\citet{bohnet-etal-2023-coreference} and all
large-sized models not using additional training data. Unfortunately, we did
not have the resources to train an xxl-sized model.

\section{Conclusions}

We presented CorPipe 24, the winning entry to the \CRAClong~\cite{novak-etal-2024-findings}.
Our system has two variants, either first predicting empty nodes using
a pretrained language encoder model and then performing coreference resolution
employing another pretrained model, or predicting the empty nodes jointly with
mention detection and coreference linking. Both variants surpass other
participants by a large margin of 3.9 and 2.8 percent points, respectively.
The source code and the trained model are available at
{\small\url{https://github.com/ufal/crac2024-corpipe}}.

\section*{Acknowledgements}

This work has been supported by the Grant Agency of the Czech Republic,
project EXPRO LUSyD (GX20-16819X), and has been using data provided by
the LINDAT/CLARIAH-CZ Research Infrastructure
({\footnotesize\url{https://lindat.cz}}) of the Ministry of Education,
Youth and Sports of the Czech Republic (Project No. LM2023062).

\section*{Limitations}
The presented system has demonstrated its performance only on a limited set of
15 languages, and heavily depends on a large pretrained model, transitively
receiving its limitations and biases.

Training with the mT5-large pretrained model requires a 40GB GPU, which we
consider affordable; however, training with the mT5-xl pretrained model needs
nearly four times as much GPU memory.

\bigbreak

\bibliography{anthology,custom}

\end{document}

%% file: table_empty_nodes_baseline.tex
  \setlength{\tabcolsep}{3pt}
  \begin{tabular*}{\hsize}{l*{3}{@{\extracolsep{\fill}}c}}
    \toprule
      Treebank & Precison & Recall & $F_1$-score \\
    \midrule
      \texttt{ca}& 92.32 & 91.01 & 91.66 \\
      \texttt{cs\_pcedt} & 78.22 & 59.84 & 67.81 \\
      \texttt{cs\_pdt} & 81.47 & 71.56 & 76.19 \\
      \texttt{cu} & 81.61 & 78.76 & 80.16 \\
      \texttt{es} & 92.04 & 91.92 & 91.98 \\
      \texttt{grc} & 90.29 & 86.58 & 88.39 \\
      \texttt{hu\_korkor} & 74.68 & 60.21 & 66.67 \\
      \texttt{hu\_szegedkoref} & 91.93 & 89.52 & 90.71 \\
      \texttt{pl} & 87.50 & 91.61 & 89.51 \\
      \texttt{tr} & 79.05 & 93.81 & 85.80 \\
    \bottomrule
  \end{tabular*}

%% file: table_official_treebanks.tex
  \scriptsize\setlength{\tabcolsep}{3pt}
  \renewcommand\cellset{\renewcommand\arraystretch{0.85}}\renewcommand{\arraystretch}{1.95}
  \catcode`@ = 13\def@{\bfseries}
  \begin{tabular*}{\hsize}{l*{22}{@{\extracolsep{\fill}}c}}
    \toprule
      System & Avg &
      \texttt{ca} &
      \makecell[c]{\texttt{cs} \\ \texttt{\kern-.1em pced\kern-.1em}} &
      \makecell[c]{\texttt{cs} \\ \texttt{pdt}} &
      \texttt{cu} &
      \makecell[c]{\texttt{de} \\ \texttt{\kern-.1em parc\kern-.1em}} &
      \makecell[c]{\texttt{de} \\ \texttt{\kern-.1em pots\kern-.1em}} &
      \makecell[c]{\texttt{en} \\ \texttt{gum}} &
      \makecell[c]{\texttt{en} \\ \texttt{\kern-.1em litb\kern-.1em}} &
      \makecell[c]{\texttt{en} \\ \texttt{\kern-.1em parc\kern-.1em}} &
      \texttt{es} &
      \texttt{fr} &
      \texttt{grc} &
      \texttt{hbo} &
      \makecell[c]{\texttt{hu} \\ \texttt{\kern-.1em kork\kern-.1em}} &
      \makecell[c]{\texttt{hu} \\ \texttt{\kern-.1em szeg\kern-.1em}} &
      \texttt{lt} &
      \makecell[c]{\texttt{no} \\ \texttt{\kern-.1em bokm\kern-.1em}} &
      \makecell[c]{\texttt{no} \\ \texttt{\kern-.1em nyno\kern-.1em}} &
      \texttt{pl} &
      \texttt{ru} &
      \texttt{tr} \\
  \midrule
    @CorPipe-2stage &@\makecell[c]{73.90 \\ 1} & \makecell[c]{82.2 \\ 2} &@\makecell[c]{74.8 \\ 1} &@\makecell[c]{77.2 \\ 1} &@\makecell[c]{61.6 \\ 1} & \makecell[c]{69.5 \\ 3} & \makecell[c]{71.8 \\ 2} &@\makecell[c]{75.7 \\ 1} &@\makecell[c]{79.6 \\ 1} & \makecell[c]{68.9 \\ 2} &@\makecell[c]{82.5 \\ 1} & \makecell[c]{68.2 \\ 2} &@\makecell[c]{71.3 \\ 1} &@\makecell[c]{72.0 \\ 1} & \makecell[c]{63.2 \\ 2} &@\makecell[c]{70.0 \\ 1} &@\makecell[c]{75.8 \\ 1} &@\makecell[c]{79.8 \\ 1} &@\makecell[c]{78.0 \\ 1} &@\makecell[c]{78.5 \\ 1} &@\makecell[c]{83.2 \\ 1} &@\makecell[c]{68.2 \\ 1} \\
  @CorPipe & \makecell[c]{72.75 \\ 2} & \makecell[c]{81.0 \\ 3} & \makecell[c]{73.7 \\ 2} & \makecell[c]{75.8 \\ 2} & \makecell[c]{60.7 \\ 2} &@\makecell[c]{71.7 \\ 1} & \makecell[c]{71.5 \\ 3} & \makecell[c]{74.6 \\ 2} & \makecell[c]{79.1 \\ 2} &@\makecell[c]{69.8 \\ 1} & \makecell[c]{81.0 \\ 3} &@\makecell[c]{68.8 \\ 1} & \makecell[c]{68.5 \\ 2} & \makecell[c]{70.9 \\ 2} & \makecell[c]{60.3 \\ 3} & \makecell[c]{68.1 \\ 3} & \makecell[c]{75.8 \\ 2} & \makecell[c]{79.5 \\ 2} & \makecell[c]{77.5 \\ 2} & \makecell[c]{77.0 \\ 2} & \makecell[c]{83.1 \\ 2} & \makecell[c]{59.4 \\ 3} \\
  @CorPipe-single & \makecell[c]{70.18 \\ 3} & \makecell[c]{80.4 \\ 4} & \makecell[c]{72.8 \\ 3} & \makecell[c]{74.8 \\ 4} & \makecell[c]{57.1 \\ 3} & \makecell[c]{61.6 \\ 4} & \makecell[c]{67.0 \\ 4} & \makecell[c]{74.4 \\ 3} & \makecell[c]{78.1 \\ 3} & \makecell[c]{58.6 \\ 3} & \makecell[c]{79.8 \\ 4} & \makecell[c]{67.9 \\ 3} & \makecell[c]{66.0 \\ 3} & \makecell[c]{67.2 \\ 3} & \makecell[c]{60.1 \\ 4} & \makecell[c]{67.3 \\ 4} & \makecell[c]{75.2 \\ 3} & \makecell[c]{78.9 \\ 3} & \makecell[c]{76.6 \\ 3} & \makecell[c]{75.2 \\ 4} & \makecell[c]{81.2 \\ 3} & \makecell[c]{53.4 \\ 4} \\
  Ondfa & \makecell[c]{69.97 \\ 4} &@\makecell[c]{82.5 \\ 1} & \makecell[c]{70.8 \\ 4} & \makecell[c]{75.8 \\ 3} & \makecell[c]{55.0 \\ 4} & \makecell[c]{71.4 \\ 2} &@\makecell[c]{71.9 \\ 1} & \makecell[c]{70.5 \\ 4} & \makecell[c]{74.2 \\ 4} & \makecell[c]{55.6 \\ 4} & \makecell[c]{81.9 \\ 2} & \makecell[c]{62.7 \\ 4} & \makecell[c]{61.6 \\ 4} & \makecell[c]{61.6 \\ 4} &@\makecell[c]{64.9 \\ 1} & \makecell[c]{69.3 \\ 2} & \makecell[c]{72.0 \\ 4} & \makecell[c]{74.5 \\ 4} & \makecell[c]{72.1 \\ 4} & \makecell[c]{76.3 \\ 3} & \makecell[c]{80.5 \\ 4} & \makecell[c]{64.5 \\ 2} \\
  BASELINE$^\dagger$ & \makecell[c]{53.16 \\ 5} & \makecell[c]{68.3 \\ 5} & \makecell[c]{64.1 \\ 5} & \makecell[c]{63.8 \\ 5} & \makecell[c]{24.5 \\ 5} & \makecell[c]{47.2 \\ 5} & \makecell[c]{55.6 \\ 5} & \makecell[c]{63.2 \\ 5} & \makecell[c]{63.5 \\ 5} & \makecell[c]{33.1 \\ 6} & \makecell[c]{69.6 \\ 5} & \makecell[c]{53.6 \\ 5} & \makecell[c]{28.8 \\ 5} & \makecell[c]{24.6 \\ 6} & \makecell[c]{35.1 \\ 5} & \makecell[c]{54.5 \\ 5} & \makecell[c]{62.0 \\ 5} & \makecell[c]{65.0 \\ 5} & \makecell[c]{63.7 \\ 5} & \makecell[c]{66.2 \\ 5} & \makecell[c]{65.8 \\ 5} & \makecell[c]{44.0 \\ 5} \\
  DFKI-CorefGen & \makecell[c]{33.38 \\ 6} & \makecell[c]{34.8 \\ 6} & \makecell[c]{32.9 \\ 6} & \makecell[c]{30.9 \\ 6} & \makecell[c]{22.5 \\ 6} & \makecell[c]{23.1 \\ 7} & \makecell[c]{45.9 \\ 7} & \makecell[c]{35.5 \\ 6} & \makecell[c]{46.6 \\ 6} & \makecell[c]{32.7 \\ 7} & \makecell[c]{37.8 \\ 6} & \makecell[c]{36.3 \\ 7} & \makecell[c]{25.9 \\ 6} & \makecell[c]{38.0 \\ 5} & \makecell[c]{23.5 \\ 7} & \makecell[c]{33.9 \\ 6} & \makecell[c]{42.7 \\ 7} & \makecell[c]{37.9 \\ 6} & \makecell[c]{35.7 \\ 6} & \makecell[c]{27.2 \\ 6} & \makecell[c]{47.8 \\ 7} & \makecell[c]{9.7 \\ 6} \\
  Ritwikmishra & \makecell[c]{16.47 \\ 7} & \makecell[c]{0.0 \\ 7} & \makecell[c]{0.0 \\ 7} & \makecell[c]{0.0 \\ 7} & \makecell[c]{6.8 \\ 7} & \makecell[c]{25.4 \\ 6} & \makecell[c]{48.9 \\ 6} & \makecell[c]{0.0 \\ 7} & \makecell[c]{0.0 \\ 7} & \makecell[c]{53.1 \\ 5} & \makecell[c]{0.0 \\ 7} & \makecell[c]{43.7 \\ 6} & \makecell[c]{5.6 \\ 7} & \makecell[c]{0.1 \\ 7} & \makecell[c]{33.4 \\ 6} & \makecell[c]{30.3 \\ 7} & \makecell[c]{44.8 \\ 6} & \makecell[c]{0.0 \\ 7} & \makecell[c]{0.0 \\ 7} & \makecell[c]{0.0 \\ 7} & \makecell[c]{53.9 \\ 6} & \makecell[c]{0.0 \\ 7} \\
    \bottomrule
\end{tabular*}

%% file: table_official_metrics.tex
  \footnotesize\setlength{\tabcolsep}{3pt}
  \renewcommand\cellset{\renewcommand\arraystretch{0.85}}\renewcommand{\arraystretch}{1.95}
  \catcode`@ = 13\def@{\bfseries}
  \begin{tabular*}{\hsize}{l*{4}{@{\extracolsep{\fill}}c}}
    \toprule
      System & \makecell{Head-\\match} & \makecell{Partial-\\match} & \makecell{Exact-\\match} & \makecell{With Sin-\\gletons} \\
  \midrule
    @CorPipe-2stage &@\makecell[c]{73.90 \\ 1} &@\makecell[c]{72.19 \\ 1} &@\makecell[c]{69.86 \\ 1} &@\makecell[c]{75.65 \\ 1} \\
    @CorPipe        & \makecell[c]{72.75 \\ 2} & \makecell[c]{70.30 \\ 2} & \makecell[c]{68.36 \\ 2} & \makecell[c]{74.65 \\ 2} \\
    @CorPipe-single & \makecell[c]{70.18 \\ 3} & \makecell[c]{68.02 \\ 4} & \makecell[c]{66.07 \\ 3} & \makecell[c]{71.96 \\ 3} \\
    Ondfa           & \makecell[c]{69.97 \\ 4} & \makecell[c]{69.82 \\ 3} & \makecell[c]{40.25 \\ 5} & \makecell[c]{70.67 \\ 4} \\
    BASELINE        & \makecell[c]{53.16 \\ 5} & \makecell[c]{52.48 \\ 5} & \makecell[c]{51.26 \\ 4} & \makecell[c]{46.45 \\ 5} \\
    DFKI-CorefGen   & \makecell[c]{33.38 \\ 6} & \makecell[c]{32.36 \\ 6} & \makecell[c]{30.71 \\ 6} & \makecell[c]{38.65 \\ 6} \\
    Ritwikmishra    & \makecell[c]{16.47 \\ 7} & \makecell[c]{16.65 \\ 7} & \makecell[c]{14.16 \\ 7} & \makecell[c]{15.42 \\ 7} \\
  \bottomrule
\end{tabular*}

%% file: table_test_results.tex
    \scriptsize\setlength{\tabcolsep}{1pt}
    \catcode`@ = 13\def@{\bfseries}
    \catcode`! = 13\def!{\itshape}
    \begin{tabular*}{\hsize}{l*{22}{@{\extracolsep{\fill}}c}}
      \toprule
        System & Avg &
        \texttt{ca} &
        \makecell[c]{\texttt{cs} \\ \texttt{\kern-.1em pced\kern-.1em}} &
        \makecell[c]{\texttt{cs} \\ \texttt{pdt}} &
        \texttt{cu} &
        \makecell[c]{\texttt{de} \\ \texttt{\kern-.1em parc\kern-.1em}} &
        \makecell[c]{\texttt{de} \\ \texttt{\kern-.1em pots\kern-.1em}} &
        \makecell[c]{\texttt{en} \\ \texttt{gum}} &
        \makecell[c]{\texttt{en} \\ \texttt{\kern-.1em litb\kern-.1em}} &
        \makecell[c]{\texttt{en} \\ \texttt{\kern-.1em parc\kern-.1em}} &
        \texttt{es} &
        \texttt{fr} &
        \texttt{grc} &
        \texttt{hbo} &
        \makecell[c]{\texttt{hu} \\ \texttt{\kern-.1em kork\kern-.1em}} &
        \makecell[c]{\texttt{hu} \\ \texttt{\kern-.1em szeg\kern-.1em}} &
        \texttt{lt} &
        \makecell[c]{\texttt{no} \\ \texttt{\kern-.1em bokm\kern-.1em}} &
        \makecell[c]{\texttt{no} \\ \texttt{\kern-.1em nyno\kern-.1em}} &
        \texttt{pl} &
        \texttt{ru} &
        \texttt{tr} \\
    \midrule
\multicolumn{22}{l}{\textsc{A) CorPipe Single-stage Variants}} \\[2pt]
~~Single model & \textcolor{black}{70.18} & \textcolor{black}{80.4} & \textcolor{black}{72.8} & \textcolor{black}{74.8} & \textcolor{black}{57.1} & \textcolor{black}{61.6} & \textcolor{black}{67.0} & \textcolor{black}{74.4} & \textcolor{black}{78.1} & \textcolor{black}{58.6} & \textcolor{black}{79.8} & \textcolor{black}{67.9} & \textcolor{black}{66.0} & \textcolor{black}{67.2} & \textcolor{black}{60.1} & \textcolor{black}{67.3} & \textcolor{black}{75.2} & \textcolor{black}{78.9} & \textcolor{black}{76.6} & \textcolor{black}{75.2} & \textcolor{black}{81.2} & \textcolor{black}{53.4} \\
~~Per-corpus best & \textcolor{blue!54.9!black}{+1.42} & \textcolor{red!100.0!black}{--\kern 0.04em 0.4} & \textcolor{red!100.0!black}{--\kern 0.04em 0.6} & \textcolor{red!100.0!black}{--\kern 0.04em 0.2} & \textcolor{blue!70.1!black}{+2.5} & \textcolor{blue!71.8!black}{+7.2} & \textcolor{blue!60.9!black}{+2.7} & \textcolor{red!100.0!black}{--\kern 0.04em 0.4} & \textcolor{red!100.0!black}{--\kern 0.04em 0.6} & \textcolor{blue!93.6!black}{+10.4} & \textcolor{red!100.0!black}{--\kern 0.04em 0.0} & \textcolor{red!100.0!black}{--\kern 0.04em 0.3} & \textcolor{blue!38.9!black}{+1.0} & \textcolor{blue!41.3!black}{+1.5} & \textcolor{blue!100.0!black}{@+2.5} & \textcolor{red!100.0!black}{--\kern 0.04em 1.6} & \textcolor{blue!100.0!black}{@+0.9} & \textcolor{red!100.0!black}{--\kern 0.04em 0.4} & \textcolor{blue!92.4!black}{+0.9} & \textcolor{red!100.0!black}{--\kern 0.04em 0.2} & \textcolor{red!100.0!black}{--\kern 0.04em 0.2} & \textcolor{blue!84.5!black}{+5.1} \\
~~Per-corpus ensemble & \textcolor{blue!100.0!black}{@+2.62} & \textcolor{blue!100.0!black}{@+0.6} & \textcolor{blue!100.0!black}{@+0.9} & \textcolor{blue!100.0!black}{@+1.0} & \textcolor{blue!100.0!black}{@+3.6} & \textcolor{blue!100.0!black}{@+10.1} & \textcolor{blue!100.0!black}{@+4.5} & \textcolor{blue!100.0!black}{@+0.2} & \textcolor{blue!100.0!black}{@+1.0} & \textcolor{blue!100.0!black}{@+11.2} & \textcolor{blue!100.0!black}{@+1.2} & \textcolor{blue!100.0!black}{@+0.9} & \textcolor{blue!100.0!black}{@+2.5} & \textcolor{blue!100.0!black}{@+3.7} & \textcolor{blue!9.1!black}{+0.2} & \textcolor{blue!100.0!black}{@+0.8} & \textcolor{blue!64.8!black}{+0.6} & \textcolor{blue!100.0!black}{@+0.6} & \textcolor{blue!100.0!black}{@+0.9} & \textcolor{blue!100.0!black}{@+1.8} & \textcolor{blue!100.0!black}{@+1.9} & \textcolor{blue!100.0!black}{@+6.0} \\
    \midrule
\multicolumn{22}{l}{\textsc{B) CorPipe Two-stage Variants}} \\[2pt]
~~Single model & \textcolor{black}{71.32} & \textcolor{black}{81.0} & \textcolor{black}{74.2} & \textcolor{black}{75.9} & \textcolor{black}{56.7} & \textcolor{black}{64.7} & \textcolor{black}{66.4} & \textcolor{black}{74.7} & \textcolor{black}{78.2} & \textcolor{black}{57.9} & \textcolor{black}{81.2} & \textcolor{black}{67.2} & \textcolor{black}{67.6} & \textcolor{black}{64.2} & \textcolor{black}{61.6} & \textcolor{black}{67.9} & \textcolor{black}{@77.7} & \textcolor{black}{77.6} & \textcolor{black}{77.3} & \textcolor{black}{77.4} & \textcolor{black}{81.3} & \textcolor{black}{67.0} \\
~~Per-corpus best & \textcolor{blue!46.1!black}{+1.18} & \textcolor{blue!11.4!black}{+0.1} & \textcolor{blue!57.1!black}{+0.4} & \textcolor{blue!29.0!black}{+0.3} & \textcolor{blue!74.6!black}{+3.7} & \textcolor{blue!100.0!black}{@+4.9} & \textcolor{blue!11.5!black}{+0.6} & \textcolor{red!100.0!black}{--\kern 0.04em 1.2} & \textcolor{blue!32.4!black}{+0.5} & \textcolor{blue!92.6!black}{+10.2} & \textcolor{blue!52.8!black}{+0.7} & \textcolor{red!100.0!black}{--\kern 0.04em 0.2} & \textcolor{blue!35.8!black}{+1.3} & \textcolor{blue!71.0!black}{+5.6} & \textcolor{red!100.0!black}{--\kern 0.04em 0.2} & \textcolor{red!100.0!black}{--\kern 0.04em 0.6} & \textcolor{red!100.0!black}{--\kern 0.04em 4.2} & \textcolor{blue!97.7!black}{+2.2} & \textcolor{blue!49.3!black}{+0.4} & \textcolor{blue!45.9!black}{+0.5} & \textcolor{red!100.0!black}{--\kern 0.04em 0.1} & \textcolor{blue!12.7!black}{+0.2} \\
~~Per-corpus ensemble & \textcolor{blue!100.0!black}{@+2.58} & \textcolor{blue!100.0!black}{@+1.2} & \textcolor{blue!100.0!black}{@+0.6} & \textcolor{blue!100.0!black}{@+1.3} & \textcolor{blue!100.0!black}{@+4.9} & \textcolor{blue!98.6!black}{+4.8} & \textcolor{blue!100.0!black}{@+5.4} & \textcolor{blue!100.0!black}{@+1.0} & \textcolor{blue!100.0!black}{@+1.4} & \textcolor{blue!100.0!black}{@+11.1} & \textcolor{blue!100.0!black}{@+1.3} & \textcolor{blue!100.0!black}{@+1.0} & \textcolor{blue!100.0!black}{@+3.7} & \textcolor{blue!100.0!black}{@+7.8} & \textcolor{blue!100.0!black}{@+1.6} & \textcolor{blue!100.0!black}{@+2.1} & \textcolor{red!44.8!black}{--\kern 0.04em 1.9} & \textcolor{blue!100.0!black}{@+2.2} & \textcolor{blue!100.0!black}{@+0.7} & \textcolor{blue!100.0!black}{@+1.1} & \textcolor{blue!100.0!black}{@+1.9} & \textcolor{blue!100.0!black}{@+1.2} \\
    \midrule
\multicolumn{22}{l}{\textsc{C) Comparison of Single-model Variants}} \\[2pt]
~~Single-stage & \textcolor{black}{70.18} & \textcolor{black}{80.4} & \textcolor{black}{72.8} & \textcolor{black}{74.8} & \textcolor{black}{@57.1} & \textcolor{black}{61.6} & \textcolor{black}{@67.0} & \textcolor{black}{74.4} & \textcolor{black}{78.1} & \textcolor{black}{@58.6} & \textcolor{black}{79.8} & \textcolor{black}{@67.9} & \textcolor{black}{66.0} & \textcolor{black}{@67.2} & \textcolor{black}{60.1} & \textcolor{black}{67.3} & \textcolor{black}{75.2} & \textcolor{black}{@78.9} & \textcolor{black}{76.6} & \textcolor{black}{75.2} & \textcolor{black}{81.2} & \textcolor{black}{53.4} \\
~~Two-stage & \textcolor{blue!100.0!black}{@+1.12} & \textcolor{blue!100.0!black}{@+0.6} & \textcolor{blue!100.0!black}{@+1.4} & \textcolor{blue!100.0!black}{@+1.1} & \textcolor{red!100.0!black}{--\kern 0.04em 0.4} & \textcolor{blue!100.0!black}{@+3.1} & \textcolor{red!100.0!black}{--\kern 0.04em 0.6} & \textcolor{blue!100.0!black}{@+0.3} & \textcolor{blue!100.0!black}{@+0.1} & \textcolor{red!100.0!black}{--\kern 0.04em 0.7} & \textcolor{blue!100.0!black}{@+1.5} & \textcolor{red!100.0!black}{--\kern 0.04em 0.7} & \textcolor{blue!100.0!black}{@+1.6} & \textcolor{red!100.0!black}{--\kern 0.04em 3.0} & \textcolor{blue!100.0!black}{@+1.5} & \textcolor{blue!100.0!black}{@+0.6} & \textcolor{blue!100.0!black}{@+2.5} & \textcolor{red!100.0!black}{--\kern 0.04em 1.3} & \textcolor{blue!100.0!black}{@+0.7} & \textcolor{blue!100.0!black}{@+2.2} & \textcolor{blue!100.0!black}{@+0.1} & \textcolor{blue!100.0!black}{@+13.6} \\
    \midrule
\multicolumn{22}{l}{\textsc{D) Comparison of Per-corpus Best Variants}} \\[2pt]
~~Single-stage & \textcolor{black}{71.59} & \textcolor{black}{80.0} & \textcolor{black}{72.2} & \textcolor{black}{74.6} & \textcolor{black}{59.6} & \textcolor{black}{68.8} & \textcolor{black}{@69.7} & \textcolor{black}{@74.0} & \textcolor{black}{77.5} & \textcolor{black}{@69.0} & \textcolor{black}{79.7} & \textcolor{black}{@67.6} & \textcolor{black}{67.0} & \textcolor{black}{68.7} & \textcolor{black}{@62.6} & \textcolor{black}{65.7} & \textcolor{black}{@76.1} & \textcolor{black}{78.5} & \textcolor{black}{77.5} & \textcolor{black}{75.0} & \textcolor{black}{81.0} & \textcolor{black}{58.5} \\
~~Two-stage & \textcolor{blue!100.0!black}{@+0.91} & \textcolor{blue!100.0!black}{@+1.1} & \textcolor{blue!100.0!black}{@+2.4} & \textcolor{blue!100.0!black}{@+1.6} & \textcolor{blue!100.0!black}{@+0.8} & \textcolor{blue!100.0!black}{@+0.8} & \textcolor{red!100.0!black}{--\kern 0.04em 2.7} & \textcolor{red!100.0!black}{--\kern 0.04em 0.5} & \textcolor{blue!100.0!black}{@+1.2} & \textcolor{red!100.0!black}{--\kern 0.04em 0.9} & \textcolor{blue!100.0!black}{@+2.2} & \textcolor{red!100.0!black}{--\kern 0.04em 0.6} & \textcolor{blue!100.0!black}{@+1.9} & \textcolor{blue!100.0!black}{@+1.1} & \textcolor{red!100.0!black}{--\kern 0.04em 1.2} & \textcolor{blue!100.0!black}{@+1.6} & \textcolor{red!100.0!black}{--\kern 0.04em 2.6} & \textcolor{blue!100.0!black}{@+1.3} & \textcolor{blue!100.0!black}{@+0.2} & \textcolor{blue!100.0!black}{@+2.9} & \textcolor{blue!100.0!black}{@+0.2} & \textcolor{blue!100.0!black}{@+8.8} \\
    \midrule
\multicolumn{22}{l}{\textsc{E) Comparison of Per-corpus Ensemble Variants}} \\[2pt]
~~Single-stage & \textcolor{black}{72.75} & \textcolor{black}{81.0} & \textcolor{black}{73.7} & \textcolor{black}{75.8} & \textcolor{black}{60.7} & \textcolor{black}{@71.7} & \textcolor{black}{71.5} & \textcolor{black}{74.6} & \textcolor{black}{79.1} & \textcolor{black}{@69.8} & \textcolor{black}{81.0} & \textcolor{black}{@68.8} & \textcolor{black}{68.5} & \textcolor{black}{70.9} & \textcolor{black}{60.3} & \textcolor{black}{68.1} & \textcolor{black}{75.8} & \textcolor{black}{79.5} & \textcolor{black}{77.5} & \textcolor{black}{77.0} & \textcolor{black}{83.1} & \textcolor{black}{59.4} \\
~~Two-stage & \textcolor{blue!100.0!black}{@+1.15} & \textcolor{blue!100.0!black}{@+1.2} & \textcolor{blue!100.0!black}{@+1.1} & \textcolor{blue!100.0!black}{@+1.4} & \textcolor{blue!100.0!black}{@+0.9} & \textcolor{red!100.0!black}{--\kern 0.04em 2.2} & \textcolor{blue!100.0!black}{@+0.3} & \textcolor{blue!100.0!black}{@+1.1} & \textcolor{blue!100.0!black}{@+0.5} & \textcolor{red!100.0!black}{--\kern 0.04em 0.8} & \textcolor{blue!100.0!black}{@+1.5} & \textcolor{red!100.0!black}{--\kern 0.04em 0.6} & \textcolor{blue!100.0!black}{@+2.8} & \textcolor{blue!100.0!black}{@+1.1} & \textcolor{blue!100.0!black}{@+2.9} & \textcolor{blue!100.0!black}{@+1.9} & \textcolor{blue!100.0!black}{@+0.0} & \textcolor{blue!100.0!black}{@+0.2} & \textcolor{blue!100.0!black}{@+0.5} & \textcolor{blue!100.0!black}{@+1.5} & \textcolor{blue!100.0!black}{@+0.1} & \textcolor{blue!100.0!black}{@+8.8} \\
    \midrule
\multicolumn{22}{l}{\textsc{F) Comparison of the Baseline System with Gold and Predicted Empty Nodes}} \\[2pt]
~~!Predicted empty nodes & \textcolor{black}{53.16} & \textcolor{black}{68.3} & \textcolor{black}{64.1} & \textcolor{black}{63.8} & \textcolor{black}{24.5} & \textcolor{black}{@47.2} & \textcolor{black}{@55.6} & \textcolor{black}{@63.2} & \textcolor{black}{@63.5} & \textcolor{black}{@33.1} & \textcolor{black}{69.6} & \textcolor{black}{@53.6} & \textcolor{black}{28.8} & \textcolor{black}{@24.6} & \textcolor{black}{35.1} & \textcolor{black}{54.5} & \textcolor{black}{@62.0} & \textcolor{black}{@65.0} & \textcolor{black}{@63.7} & \textcolor{black}{66.2} & \textcolor{black}{@65.8} & \textcolor{black}{44.0} \\
~~!Gold empty nodes & \textcolor{blue!100.0!black}{@+1.44} & \textcolor{blue!100.0!black}{@+1.3} & \textcolor{blue!100.0!black}{@+4.8} & \textcolor{blue!100.0!black}{@+2.4} & \textcolor{blue!100.0!black}{@+3.1} & ~~0.0 & ~~0.0 & ~~0.0 & ~~0.0 & ~~0.0 & \textcolor{blue!100.0!black}{@+1.0} & ~~0.0 & \textcolor{blue!100.0!black}{@+3.1} & ~~0.0 & \textcolor{blue!100.0!black}{@+6.5} & \textcolor{blue!100.0!black}{@+0.1} & ~~0.0 & ~~0.0 & ~~0.0 & \textcolor{blue!100.0!black}{@+0.8} & ~~0.0 & \textcolor{blue!100.0!black}{@+7.2} \\
    \midrule
\multicolumn{22}{l}{\textsc{G) Comparison of the CorPipe-2stage Ensemble System and the CRAC23 Best Results}} \\[2pt]
~~CorPipe-2stage, ensemble & \textcolor{black}{74.55} & \textcolor{black}{82.2} & \textcolor{black}{74.8} & \textcolor{black}{77.2} & --- & \textcolor{black}{69.5} & \textcolor{black}{71.8} & \textcolor{black}{75.7} & --- & \textcolor{black}{68.9} & \textcolor{black}{82.5} & \textcolor{black}{68.2} & --- & --- & \textcolor{black}{63.2} & \textcolor{black}{70.0} & \textcolor{black}{75.8} & \textcolor{black}{@79.8} & \textcolor{black}{78.0} & \textcolor{black}{78.5} & \textcolor{black}{@83.2} & \textcolor{black}{@68.2} \\
~~!CorPipe23, CRAC23 & \textcolor{blue!100.0!black}{@+0.65} & \textcolor{blue!100.0!black}{@+1.0} & \textcolor{blue!100.0!black}{@+4.5} & \textcolor{blue!100.0!black}{@+2.3} & --- & \textcolor{blue!100.0!black}{@+1.5} & \textcolor{blue!100.0!black}{@+0.0} & \textcolor{blue!100.0!black}{@+0.8} & --- & \textcolor{blue!100.0!black}{@+2.1} & \textcolor{blue!100.0!black}{@+1.0} & \textcolor{blue!100.0!black}{@+0.4} & --- & --- & \textcolor{blue!100.0!black}{@+6.3} & \textcolor{blue!100.0!black}{@+0.8} & \textcolor{blue!100.0!black}{@+0.6} & \textcolor{red!100.0!black}{--\kern 0.04em 0.2} & \textcolor{blue!100.0!black}{@+1.0} & \textcolor{blue!100.0!black}{@+1.3} & \textcolor{red!100.0!black}{--\kern 0.04em 0.6} & \textcolor{red!100.0!black}{--\kern 0.04em 11.7} \\
    \bottomrule
    \end{tabular*}

%% file: table_ontonotes.tex
  \setlength{\tabcolsep}{3pt}
  \begin{tabular*}{\hsize}{l@{\extracolsep{\fill}}l@{\extracolsep{\fill}}c*{4}{@{\extracolsep{\fill}}l}}
    \toprule
      Paper & Model & \kern-.5em\makecell{\#model\\calls}\kern-.5em
      & \makecell{$\varnothing$, ELMO,\\base PLM}
      & \makecell{large PLM\\$\sim$350M}
      & \makecell{xl PLM\\$\sim$3B}
      & \makecell{xxl PLM\\$\sim$11B} \\
  \midrule
    \citep{lee-etal-2017-end} & e2e & 1 & 67.2$_\varnothing$ \\
    \citep{lee-etal-2018-higher} & e2e & 1 & 70.4$_\mathrm{ELMO}$ \\
    \citep{lee-etal-2018-higher} & c2f & 1 & 73.0$_\mathrm{ELMO}$ \\
    \citep{joshi-etal-2019-bert} & c2f & 1 & 73.9$_\mathrm{BERT}$ & 76.9$_\mathrm{BERT}$ \\
    \citep{joshi-etal-2020-spanbert} & c2f & 1 & & 79.6\rlap{$_\mathrm{SpanBERT}$} \\
    \citep{kirstain-etal-2021-coreference} & s2e & 1 & & 80.3\rlap{$_\mathrm{Longformer}$} \\
    \citep{otmazgin-etal-2023-lingmess} & s2e/LingM\rlap{ess} & 1 & & 81.4\rlap{$^\mathrm{+additional~annotations}_\mathrm{Longformer}$} \\
    \citep{dobrovolskii-2021-word} & WL & 1 & & 81.0\rlap{$_\mathrm{RoBERTa}$} \\
    \citep{doosterlinck-etal-2023-caw} & WL/CAW & 1 & & 81.6\rlap{$_\mathrm{RoBERTa}$} \\
  \midrule
    \citep{liu-etal-2022-autoregressive} & ASP & $\mathcal{O}(n)$ & 76.6$_\mathrm{T5}$ & 79.3$_\mathrm{T5}$ & 82.3$_\mathrm{T0}$ & 82.5$_\mathrm{FlanT5}$ \\
    \citep{bohnet-etal-2023-coreference} & seq2seq & $\mathcal{O}(n)$ & & & 78.0$^\mathrm{dev}_\mathrm{mT5}$ & 83.3$_\mathrm{mT5}$ \\
    \citep{wu-etal-2020-corefqa} & CorefQA & $\mathcal{O}(n)$ & 79.9$^\mathrm{+QA~data}_\mathrm{SpanBERT}$ & 83.1$^\mathrm{+QA~data}_\mathrm{SpanBERT}$\\
  \midrule
    This paper & CorPipe & 1 & & 80.7$_\mathrm{T5}$ & 82.0$_\mathrm{FlanT5}$ \\
    This paper & CorPipe & 1 & & 77.2$_\mathrm{mT5}$ & 78.9$_\mathrm{mT5}$ \\
  \bottomrule
\end{tabular*}